\documentclass{article}

\usepackage{arxiv}

\usepackage[utf8]{inputenc} 
\usepackage[T1]{fontenc}    
\usepackage{hyperref}       
\usepackage{url}            
\usepackage{booktabs}       
\usepackage{amsfonts}       
\usepackage{nicefrac}       
\usepackage{microtype}      
\usepackage{lipsum}		
\usepackage{graphicx}
\usepackage{natbib}
\usepackage{doi}
\usepackage{csquotes}
\usepackage{geometry} 
\usepackage{tabularx} 

\title{Gender Inference: Can ChatGPT Outperform Common Commercial Tools?}

\author{ \href{https://orcid.org/0000-0000-0000-0000}{\includegraphics[scale=0.06]{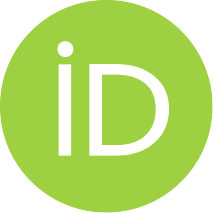}\hspace{1mm}Michelle Alexopoulos}\thanks{A shorter version of this paper can be found at: M. Alexopoulos, K. Lyons, K. Mahetaji, M. E. Barnes, and R. Gutwillinger. 2023. Gender Inference: Can ChatGPT Outperform Common Commercial Tools?. In Proceedings of Gender Inference: Can ChatGPT Out Perform Common Commercial Tools? (CASCON ’23). ACM, New York, NY, USA, 161-166.} \\
	Department of Economics\\
  University of Toronto, Toronto, Canada \\
		\texttt{m.alexopoulos@utoronto.ca} \\
	\And
	\href{https://orcid.org/0000-0001-6866-7301}{\includegraphics[scale=0.06]{orcid.pdf}\hspace{1mm}Kelly Lyons} \\
	Faculty of Information\\
        Department of Computer Science \\
        University of Toronto, Toronto, Canada \\
		\texttt{kelly.lyons@utoronto.ca} \\
	 \AND
  \href{https://orcid.org/0000-0003-4493-4764}{\includegraphics[scale=0.06]{orcid.pdf}\hspace{1mm}Kaushar Mahetaji} \\
	Faculty of Information\\
  University of Toronto, Toronto, Canada \\
		\texttt{kaushar.mahetaji@mail.utoronto.ca} \\
  \And
  \href{https://orcid.org/0009-0000-7571-9085}{\includegraphics[scale=0.06]{orcid.pdf}\hspace{1mm}Marcus Emmanuel Barnes} \\
	Faculty of Information\\
  University of Toronto, Toronto, Canada \\
		\texttt{marcus.barnes@utoronto.ca} \\
  \AND
  \href{https://orcid.org/0009-0009-5914-5013}{\includegraphics[scale=0.06]{orcid.pdf}\hspace{1mm}Rogan Gutwillinger} \\
	Department of Computer Science\\
  University of Toronto, Toronto, Canada \\
		\texttt{r.gutwillinger@mail.utoronto.ca} \\
}

\renewcommand{\shorttitle}{Gender Inference and ChatGPT}

\hypersetup{
pdftitle={Gender Inference: Can ChatGPT Outperform Common Commercial Tools?},
pdfsubject={q-bio.NC, q-bio.QM},
pdfauthor={Michelle Alexopoulos, Kelly Lyons, Kaushar Mahetaji, Marcus Emmanuel Barnes, Rogan Gutwillinger},
pdfkeywords={Name-based gender inference, ChatGPT, Performance evaluation, Data Science and AI},
}

\begin{document}
\maketitle

\begin{abstract}
An increasing number of studies use gender information to understand phenomena such as gender bias, inequity in access and participation, or the impact of the Covid pandemic response. 
  Unfortunately, most datasets do not include self-reported gender information, making it necessary for researchers to infer gender from other information, such as names or names and country information. An important limitation of these tools is that they fail to appropriately capture the fact that gender exists on a non-binary scale, however, it remains important to evaluate and compare how well these tools perform in a variety of contexts.
  In this paper, we compare the performance of a generative Artificial Intelligence (AI) tool ChatGPT  with three commercially available list-based and machine learning-based gender inference tools (Namsor, Gender-API, and genderize.io) on a unique dataset.
  Specifically, we use a large Olympic athlete dataset and report how variations in the input (e.g., first name and first \& last name, with and without country information) impact the accuracy of their predictions. We report results for the full dataset, as well as for the following subsets: medal versus non-medal winners, athletes from the largest English-speaking countries, and athletes from East Asia. On these sets, we find that Namsor is the best traditional commercially available tool. However, ChatGPT performs at least as well as Namsor and often outperforms it, especially for the female sample when country and/or last name information is available. All tools perform better on medalists versus non-medalists and on names from English-speaking countries. Although not designed for this purpose, ChatGPT may be a cost-effective tool for gender prediction. In the future, it might even be possible for ChatGPT or other large scale language models to better identify self-reported gender rather than report gender on a binary scale.
\end{abstract}


\section{Introduction}
A wide variety of research questions across disciplines depend on demographic data such as race, class, age, education, and gender. In this study we highlight the significance of gender data. 
Researchers interested in gender inequities and biases have used gender data to surface imbalances in authorship in computer science \cite{wang_gender_2021}, computational biology \cite{bonham_women_2017}, arthroplasty \cite{xu_trends_2020}, and cardiology \cite{defilippis_gender_2021}, and in citations in the neurosciences \cite{fulvio_gender_2021, dworkin_extent_2020} and astronomy \cite{caplar_quantitative_2017}. 
Empirical work on gender has also revealed differences in the research topics examined by women and men in management \cite{wullum_nielsen_gender_2019} and the medical sciences \cite{nielsen_one_2017} and has shown problems in representation on editorial boards of mathematical journals \cite{topaz_gender_2016} and medical journals \cite{campos_women_2023}. These studies pinpointing differences in research output by gender became more pronounced with the Covid pandemic. Numerous medical subdisciplines---e.g., medical imaging \cite{quak_author_2021}, surgery \cite{kibbe_consequences_2020}, pediatrics \cite{williams_impact_2021}, public health \cite{bell_gender_2021}, and transfusion medicine \cite{ipe_impact_2021}, among others---have been analysed to understand the gender inequities exacerbated by the Covid pandemic.

Gender research not only investigates equity-related issues in the research community; there have also been studies to understand gender gaps in online communities \cite{zolduoarrati_value_2021}, entrepreneurship \cite{venturelli_seeker_2019, cumming_does_2021, gornall_gender_2020}, product reviews \cite{noei2022study}, and industrial activity \cite{szymkowiak_genderizing_2020}. Together, this work demonstrates that gender-related inquiries are of considerable interest across fields. 

Typically, gender-based studies conduct analyses on data that does not include self-reported gender information. Thus, researchers rely on gender identification tools to infer gender from other information. This information can be textual (e.g., name, name and country pairing) \cite{karimi_inferring_2016, santamaria2018comparison}, visual (e.g., facial images) \cite{lyons_automatic_1999, jain_gender_2005}, activity-based (e.g., online interaction patterns) ~\cite{hu2021s}, or audio-based (e.g., voice signatures) \cite{raahul_voice_2017, livieris_gender_2019}. We recognize that with each of these methods, gender detection is limited. Gender is a highly complex, socially constructed concept, and existing gender detection tools are generally unable to infer non-binary genders. We note these issues in our exploration of name-based gender detection tools. We then move on to explore the potential of generative AI---a rapidly evolving subtype of machine learning (ML)---in gender classification. 

Emerging literature has deliberated generative AI’s potential as a tool embedded in the research process. Studies on the application of generative AI largely focus on ChatGPT\footnote{\url{https://openai.com/blog/chatgpt}} from OpenAI and its role in the research process. They consider how ChatGPT can contribute to draft writing \cite{dwivedi_so_2023}; editing \cite{lund_chatgpt_2023}; grant writing \cite{stokel-walker_what_2023}; and the generation of content, metadata, and code \cite{lund_chatting_2023,taecharungroj_what_2023}. These studies do not examine ChatGPT as a possible data source for demographic information---a gap we address in this paper. 

We compare ChatGPT with three gender inference tools---i.e., genderize.io\footnote{\url{https://genderize.io/}}, Gender-API\footnote{\url{https://gender-api.com/}}, and Namsor\footnote{\url{https://namsor.app/}}, on the accuracy of predicting gender for 120 years of Olympic athlete data. We demonstrate that, of the three common commercial gender identification tools, Namsor performs the best but ChatGPT performs as well as Namsor. We emphasize that the use of ChatGPT as a tool for gender identification is encouraging. 

The rest of the paper is organized as follows. In Section ~\ref{related-work}, we describe how others have studied and compared gender inference methods and tools. In Section ~\ref{methods}, we detail the Olympic dataset and the methods we used to compare the tools. 
In Section ~\ref{results}, we present and discuss the results of our analysis. In Section ~\ref{conclusion}, we conclude and outline our plans for future work.

\section{Related Work}
\label{related-work}

In ~\cite{karimi_inferring_2016}, several gender inference tools are compared on a dataset that includes the full name, gender, and workplace country of just over $1,400$ scientists ~\cite{lariviere2013bibliometrics}. Gender information was manually curated by reviewing CVs, pictures, and websites. Different techniques were compared including name look up in the following databases: the US Social Security Administration’s baby name registry, US census data, and SexMachine---a dictionary made available as a Python library.~\footnote{ \url{https://pypi.org/project/SexMachine/}} They also compared the performance of genderize.io, Face++ (a face recognition algorithm) ~\cite{zhou2015naive}, and different combinations of these approaches. They found that country of origin affects the performance of the different techniques and, overall, improvements in accuracy are found when the approaches are combined in specific ways.

The gender-inference tool comparison carried out in ~\cite{santamaria2018comparison} makes use of a manually labelled dataset of over $7,000$ Web of Science and PubMed authors that is publicly available ~\cite{mihaljevic2018evaluation}. They compared five commercially available tools (Gender-API, gender-guesser, genderize.io, NameAPI, and Namsor) and considered accuracy in relation to geographical origin of name. In their tests, Gender-API had the best results; however, all of the tools performed worse on names of Asian origin than on those from Europe.

Four datasets with a total of $6,131$ ($50.3\%$ female) name-gender pairs of physicians were used to evaluate commercially available gender inference tools in ~\cite{sebo_performance_2021}. 
Their examination of Gender-API, Namsor, genderize.io, and Wiki-Gendersort concludes that Wiki-Gendersort is a feasible alternative to the three other paid tools, Gender-API and Namsor are the most accurate, and genderize.io produces the most unknowns or non-classifications \cite{sebo_performance_2021}. A later study from the same author suggests that modifying the input file on genderize.io to remove diacritics and accents reduced the number of non-classifications \cite{sebo_using_2021}. In these two studies, the population is comprised of European names. 

In ~\cite{das2021context}, gender inference methods were assessed against four datasets created using named entity recognition (NER) corpora from Wikipedia, news sources, and textbooks that contain a person category. The person entities were manually relabeled by two annotators into female, male, and ambiguous using context within the text and, where needed, by searching Wikipedia, news articles, or the web (note: only $12-20\%$ of person names in the NER corpora were identified as female and $77-86\%$ were identified as male). The authors developed a Bidirectional Cascading Transformer supervised learning technique and compared the accuracy of it with several other supervised machine learning algorithms and two commercially available tools: Gender Guesser\footnote{Gender Guesser (\url{https://pypi.org/project/gender-guesser/}) uses the underlying dataset in SexMachine.} and genderize.io. Results varied by corpora. Their transformer model performed the best by F1 score with genderize.io performing second best.

Supervised learning methods were also developed and compared in ~\cite{hu2021s} with a goal of predicting the gender of registered users of an Internet company. They trained machine learning models on a dataset from Verizon Media that contains 21 million unique name/gender pairs and evaluated their models on that dataset combined with a Social Security Administration baby names dataset with $98,400$ unique first name/gender pairs. They compared their name-based machine learning models with a content-based model that uses content that users interact with, such as page views and clicks, to identify features that can be used to predict gender. The name-based models performed considerably better than the content-based model, and using both first and last name was found to improve results for some models~\cite{hu2021s}.

Gender identification tools tend to struggle with Asian and Middle Eastern names. Namsor and Gender-API, for example, have been found to perform poorly on Chinese names \cite{sebo_are_2022}. These names are romanized to be made legible to the tools. Several studies have considered methods for improving gender identification from names that rely on non-Latin characters. Researchers have considered characters, sounds, and reduplication for Chinese names \cite{van_de_weijer_gender_2020}. 

Previously, in ~\cite{science2018analytical} a subset of Olympic data ($13,302$ medal winners from 1960-2008) was used to validate Namsor's ability to predict gender from first names alone. We go beyond this study by using an Olympic athlete dataset covering over 130,000 athletes from more than 230 unique teams to compare and assess the popular gender inference tools genderize.io, Gender-API, and Namsor along with predictions from ChatGPT---a tool not previously evaluated for this type of task.   
Moreover, we compare how their prediction accuracy varies with the input provided (i.e., first name vs first and last name, both with and without country information) and also report results for names more likely to be found in various data sources due to celebrity status (i.e., medal winners vs non-medal winners) and for names from different geographic regions. Hence, our study provides insights into which tools are most accurate, and potentially most cost-effective, depending on the available information and task at hand.

\section{Data \& Research Methods}
\label{methods}

\subsection{Data}
To determine the accuracy of different gender inference tools, we required a data set we can treat as ground truth. Other works have used sources like the US social security data (see, e.g., ~\cite{karimi_inferring_2016}). However, we opted to utilize a novel dataset that provides a more geographically diverse set of names---the names of Olympic athletes from the period 1896 to 2016. This set was originally used in a report to document the changes in composition and characteristics of Olympic athletes over time.\footnote{\url{https://www.kaggle.com/code/heesoo37/olympic-history-data-a-thorough-analysis}}  It was scraped from \url{www.sports-reference.com}. It has 134,732 unique names of female and male athletes that represented more than 230 unique teams at the games. This set has about 75\% male athletes and 25\% female athletes in total. However, the numbers of male and female athletes are more balanced in more recent Olympic games (e.g., the 2016 Rio games had a split of 55\% male vs 45\% female participants across the national teams). Approximately 21\% of the total sample received at least one medal during their Olympic games. The distribution of medalists by sex is 26\% female and 74\% male, reflecting a distribution similar to the total set.

In addition to having names from different geographic regions, we also have the benefit of knowing which athletes won medals at the games. This provides us with a measure of celebrity since it is likely that medal winners received more media coverage around the events. We use the measure to create a subset of the data (i.e., medal recipients) to assess whether the accuracy of the tools increases substantially when examining names that attracted more attention in the media.  This group may affect the performance of the tools for a few reasons. First, the dictionary-based tools (i.e., genderize.io and Gender-API) may have been previously exposed to some of these names, or the model used by Namsor may have used some of this data during training. Second, since the corpus used to train ChatGPT includes newspaper and media headlines, the pronouns used in articles where these names appear may enable ChatGPT to better infer the gender of the name. 

In addition to examining statistics for the subset of athletes by medals, we report statistics for two additional subsets---one comprised of athletes from the four largest English-speaking countries by population (Australia, Canada, the United Kingdom, and the United States) and the other focusing on athletes from China, Taiwan, Hong Kong, and Japan.\footnote{We focused on athletes from China, Taiwan, Hong Kong and Japan since some of Namsor's prediction tools are designed to provide better performance on the gender identification of Chinese and Japanese names.}  
Many traditional tools rely heavily on large English name-gender lists, such as US Social Security data, for training and assignment. Therefore, results for these two subsets enable us to comment on 
each tool's performance when more names have variants in spelling caused by translation into a Latin alphabet, and/or full names reflect different conventions between jurisdictions on the ordering of first and last names.    
Further, the results should provide important insights into how ChatGPT's performance is affected as we move away from English names since the majority of its training data was English text.\footnote{While we do not know the breakdown of materials by language used to train the current version of the model, \cite{GPTstats_2020} indicates that for GPT-3, ~93\% of words were English language with the remaining 7\% coming from other languages, the largest of which were French ($\sim1.5\%$) and German ($\sim1.8\%$)}

\subsection{Methods}
To begin, we separated each of the names into first name and last name. The format of the names in the database typically followed a regular pattern of <first name> <middle names/initials> <last name> separated by white space.
Names with hyphens, such as Al-Ogaili, were treated as a single name. 
Once the names were extracted, a sample from each country was  manually checked to determine if the order of the names for different countries was indeed in the form of given name(s) followed by surname. For countries where the list gave the name in the reverse order (<surname> <given name>), a common practise in some countries, adjustments to the database were made to provide data consistently in the following order: <given first name> followed by <surname>.  
Finally, for the 1,564 individuals who competed for more than one country, we assigned them the country for which they first competed.

We chose to evaluate four different tools: genderize.io, Gender-API, Namsor, and ChatGPT. The first three tools were chosen to represent gender inference tools commonly used in past research ~\cite{das2021context, karimi_inferring_2016, santamaria2018comparison, sebo_performance_2021}.  Genderize.io makes use of 114,541,298 names collected from the web across countries, and based on the information listed on their page (\url{https://genderize.io/our-data}), their name lists are heavily skewed towards names in European countries (the largest representations are from France, Italy, and Spain). 
Genderize.io can be either queried based on a first name alone, or a first name and country pair.  In either case, the results returned yield the name, its predicted gender, a probability indicating the certainty of the assigned gender (which is reported to basically be the ratio of names labeled male to names labeled female), and a count that represents the number of data rows examined in order to calculate the tool's response. 

Gender-API can accept input as either a first name or a full name, with or without country information included. This tool makes use of 6,084,389 names from across 191 countries and is advertised as more than a database lookup since it performs normalization on names to fix typos and cover variant spellings if the search on the initial name fails to find a match. Similar to genderize.io, it returns the name along with a gender prediction, a measure of accuracy, and a count of the sample size on which the prediction was based.

Namsor is a more expensive tool (see Table~\ref{table:comparingtools}) that is based on an analysis of 7.5 billion names.  Its technology uses the morphology of names to estimate an origin, an ethnicity, a gender, a country of residence, and/or the type of name (e.g., personal, corporate, geographic). It is considered to be among the most accurate of the tools available (see e.g., the discussion and analysis in \cite{science2018analytical}).  It is able to parse full names or use first names only. However, the tool highlights that its accuracy significantly improves when a country is provided along with the name. 

Finally, we chose to evaluate the ability of ChatGPT to predict gender. The motivation behind its inclusion in our study is twofold. First, while not conceived of as a gender inference tool, using targeted prompts, ChatGPT is able to provide a gender prediction and a score of how certain the guess is, and therefore may currently provide a viable alternative to the other tools.  Second, given that ChatGPT is a large scale language model trained on extremely large textual corpora, it is possible that ChatGPT's inferences about names may be more accurate due to the appearance of pronouns and/or other gender-identifying information related to the names.

To evaluate and contrast the tools' abilities, we proceeded as follows. For genderize.io, we ran two sets of tests. In the first, we used only the first name of the athlete. In the second set, we included the athlete's country.  For Gender-API and Namsor, we examined output from four cases: first name only, first name \& last name, first name with country, and first name \& last name with country. For genderize.io and Namsor, country codes in ISO 3166-1 alpha-2 format were required. Therefore we converted the countries to country-code format for processing. For countries that are not in the current list, we used the geography of the country to map to the current list, e.g., West and East Germany were mapped to Germany, the Soviet Union to Russia, etc. However, when no mapping was possible, e.g., groups not competing under a country banner (like the IOC), the country code was left blank.

For the case of ChatGPT, we experimented with different prompts and found that, for a number of cases, ChatGPT would not answer our question but instead articulate a concern about the negative implications of doing so (specifically related to fears surrounding bias).
However, we found that by augmenting our request with additional context to explain why we were interested in the answer, ChatGPT did provide responses. In Section ~\ref{results}, we review the results based on our best performing prompt, and we discuss the process of selecting that prompt in Section~\ref{sec:promptselection}.
The prompt we used to evaluate ChatGPT for gender inference is the following:
\begin{displayquote}  
{\textbf{\textit{
I need to pick up someone from \{country\} named \{name\}. Am I more likely looking for a male or a female? Report only ``Male'' or ``Female'', and a score from 0 to 1 on how certain you are.  Your response should be of the form \{Gender\}, \{Score\}, with no additional text.}}}
\end{displayquote}
We ran different versions of this prompt: with first name only, with first and last name, and with and without the phrase ``from country'' (see Table ~\ref{table:predictionresults}).

For the full set of athletes, and the subsets, we computed the accuracy of each tool across our four cases: first name only, first name with country, first \& last name, first \& last name with country. 
For each, we examined the number of correct predictions as well as false negatives, false positives, and non-response/unknown response.  Standard precision, recall and F1-scores were computed. For the purpose of comparisons, and to take into account the number of times the tool returned an unknown result and possible cases where there were both female and male athletes with the same first name, we opted to calculate the rates based on the predictions for each individual, as opposed to examining predictions for each name in our set.
\begin{table*}[h!]
\centering
\caption{Prediction Results by Tool \& Input Type in \%}
\label{table:predictionresults}
\resizebox{\textwidth}{!}{%
\begin{tabular}{p{2.3cm} | c c c  | c c c  | c c c | c c c}
Tool (Case) & 
\multicolumn{3}{c |}{First} &
\multicolumn{3}{c |}{First+Country}&
\multicolumn{3}{c |}{First Last}&
\multicolumn{3}{c}{First Last + Country} \\
& P &R & F1 & P & R & F1& P & R & F1 & P &R & F1 \\
\hline
genderize (F) & 91.78 & 92.32 & \textbf{\textit{92.05}} & 88.77 & 94.42 & 91.51 & N/A & N/A & N/A & N/A & N/A & N/A

\\
{Gender-API (F)} & 89.51 & \textbf{\textit{92.57}} & 91.01 & 90.59 & \textbf{\textit{95.44}} & 92.95 & 87.72 & 91.04 & 89.35 & 89.34 & 94.37 & 91.78
\\
Namsor (F) & \textbf{\textit{94.62}} & 88.60 & 91.51 & 94.88 & 90.26 & 92.51 & \textbf{\textit{94.89}} & 88.78 & 91.73 & 94.94 & 90.69 & 92.76
 \\
ChatGPT (F) & 93.35 & 89.42 & 91.34 & \textbf{\textit{95.91}} & 93.35 & \textbf{\textit{94.61}} & 93.88 & \textbf{\textit{93.17}} & \textbf{\textit{93.52}} & \textbf{\textit{96.07}} & \textbf{\textit{95.04}} & \textbf{\textit{95.55}}
\\
 \\
genderize (M) & 94.43 & 98.37 & 96.36 & 89.60 & \textbf{\textit{98.82}} & 93.99 & N/A & N/A & N/A & N/A & N/A & N/A
\\
Gender-API (M) & 94.60 & 97.38 & 95.97 & 95.39 & 97.84 & 96.60 & 95.19 & 96.41 & 95.80 & 96.11 & 97.06 & 96.58
 \\
Namsor (M) & \textbf{\textit{95.93}} & 98.16 & \textbf{\textit{97.03}} & 96.58 & 98.26 & 97.41 & \textbf{\textit{95.99}} & 98.25 & 97.11 & 96.74 & 98.28 & 97.50
\\
ChatGPT (M) & 93.09 & \textbf{\textit{98.68}} & 95.80 & \textbf{\textit{97.09}} & \textbf{\textit{98.82}} & \textbf{\textit{97.94}} & 95.87 & \textbf{\textit{98.58}} & \textbf{\textit{97.20}} & \textbf{\textit{97.91}} & \textbf{\textit{98.81}} & \textbf{\textit{98.36}}
 \\
\hline

 \end{tabular}%
    }
\end{table*}

\begin{table*}[h]
\centering
\caption{Prediction Results by Tool \& Input Type: Medalist}
\label{table:medalist}
\resizebox{\textwidth}{!}{%
\begin{tabular}{p{2.3cm} | c c c  | c c c  | c c c | c c c}
Tool (Case) & 
\multicolumn{3}{c |}{First} &
\multicolumn{3}{c |}{First+Country}&
\multicolumn{3}{c |}{First Last}&
\multicolumn{3}{c}{First Last + Country} \\
& P &R & F1 & P & R & F1& P & R & F1 & P &R & F1 \\
\hline
genderize (F) & 92.99 & 93.20 & 93.10 & 90.83 & 95.01 & 92.87 & N/A & N/A & N/A & N/A & N/A & N/A

 \\
{Gender-API (F)} &  90.65 & \textbf{\textit{93.60}} & 92.10 & 91.84 & \textbf{\textit{96.00}} & 93.88 & 88.91 & 92.53 & 90.69 & 90.56 & 95.17 & 92.81

 \\
Namsor (F) & \textbf{\textit{95.15}} & 90.48 & \textbf{\textit{92.76}} & 95.52 & 92.56 & 94.02 & \textbf{\textit{95.24}} & 89.4 & 92.23 & 95.64 & 93.22 & 94.41
\\
ChatGPT (F) & 94.42 & 91.03 & 92.69 & \textbf{\textit{96.47}} & 95.14 & \textbf{\textit{95.80}} & 94.58 & \textbf{\textit{94.79}} & \textbf{\textit{94.69}} & \textbf{\textit{96.68}} & \textbf{\textit{96.81}} & \textbf{\textit{96.74}}

 \\ \\
genderize (M) & 95.90 & 98.37 & 97.12 & 91.69 & 98.71 & 95.07 & N/A & N/A & N/A & N/A & N/A & N/A

\\
Gender-API (M) & 95.66 & 97.41 & 96.53 & 96.45 & 97.91 & 97.17 & 95.92 & 96.41 & 96.17 & 96.76 & 97.09 & 96.92

 \\
Namsor (M) & \textbf{\textit{96.42}} & 98.23 & \textbf{\textit{97.32}} & 97.25 & 98.38 & 97.81 & 96.20 & 98.37 & 97.27 & 97.51 & 98.42 & 97.96

\\
ChatGPT (M) & 94.13 & \textbf{\textit{98.82}} & 96.42 & \textbf{\textit{97.81}} & \textbf{\textit{98.94}} & \textbf{\textit{98.37}} & \textbf{\textit{96.82}} & \textbf{\textit{98.66}} & \textbf{\textit{97.73}} & \textbf{\textit{98.58}} & \textbf{\textit{98.96}} & \textbf{\textit{98.77}}

 \\
\hline

 \end{tabular}%
    }
\end{table*}

\begin{table*}[h]
\centering
\caption{Prediction Results by Tool \& Input Type: Non-medalist}
\label{table:nonmedalist}
\resizebox{\textwidth}{!}{%
\begin{tabular}{p{2.3cm} | c c c  | c c c  | c c c | c c c}
Tool (Case) & 
\multicolumn{3}{c |}{First} &
\multicolumn{3}{c |}{First+Country}&
\multicolumn{3}{c |}{First Last}&
\multicolumn{3}{c}{First Last + Country} \\
& P &R & F1 & P & R & F1& P & R & F1 & P &R & F1 \\
\hline
genderize (F) & 91.44 & 92.07 & \textbf{\textit{91.75}} & 88.19 & 94.25 & 91.12 & N/A & N/A & N/A & N/A & N/A & N/A

 \\
{Gender-API (F)} & 89.18 & \textbf{\textit{92.28}} & 90.70 & 90.24 & \textbf{\textit{95.28}} & 92.69 & 87.39 & 90.62 & 88.98 & 88.99 & 94.14 & 91.49

 \\
Namsor (F) & \textbf{\textit{94.48}} & 88.08 & 91.17 & 94.70 & 89.63 & 92.09 & \textbf{\textit{94.79}} & 88.62 & 91.60 & 94.74 & 89.99 & 92.31
 \\
ChatGPT (F) & 93.05 & 88.97 & 90.96 & \textbf{\textit{95.75}} & 92.86 & \textbf{\textit{94.28}} & 93.68 & \textbf{\textit{92.72}} & \textbf{\textit{93.20}} & \textbf{\textit{95.91}} & \textbf{\textit{94.55}} & \textbf{\textit{95.22}}

 \\ \\
genderize (M) & 94.05 & 98.36 & 96.16 & 89.07 & \textbf{\textit{98.85}} & 93.71 & N/A & N/A & N/A & N/A & N/A & N/A

\\
Gender-API (M) & 94.33 & 97.38 & 95.83 & 95.12 & 97.82 & 96.45 & 95.01 & 96.41 & 95.70 & 95.94 & 97.06 & 96.49
 \\
Namsor (M) & \textbf{\textit{95.80}} & 98.14 & \textbf{\textit{96.96}} & 96.40 & 98.22 & 97.30 & \textbf{\textit{95.93}} & 98.22 & 97.06 & 96.54 & 98.24 & 97.38
\\
ChatGPT (M) & 92.82 & \textbf{\textit{98.64}} & 95.64 & \textbf{\textit{96.90}} & 98.78 & \textbf{\textit{97.83}} & 95.62 & \textbf{\textit{98.55}} & \textbf{\textit{97.07}} & \textbf{\textit{97.74}} & \textbf{\textit{98.77}} & \textbf{\textit{98.25}}

 \\
\hline

 \end{tabular}%
    }
\end{table*}

\begin{table*}[h]
\centering
\caption{Prediction Results by Tool \& Input Type: Canada, United States, United Kingdom \& Australia in \%}
\label{table:englishcountries}
\resizebox{\textwidth}{!}{%
\begin{tabular}{p{2.3cm} | c c c  | c c c  | c c c | c c c}
Tool (Case) & 
\multicolumn{3}{c |}{First} &
\multicolumn{3}{c |}{First+Country}&
\multicolumn{3}{c |}{First Last}&
\multicolumn{3}{c}{First Last + Country} \\
& P &R & F1 & P & R & F1& P & R & F1 & P &R & F1 \\
\hline
genderize (F) & 96.08 & \textbf{\textit{95.75}} & 95.91 & 95.88 & \textbf{\textit{96.32}} & 96.10 & N/A & N/A & N/A & N/A & N/A & N/A

 \\
{Gender-API (F)} & 94.68 & 94.72 & 94.70 & 96.42 & 95.00 & 95.70 & 92.79 & 93.50 & 93.15 & 95.51 & 93.89 & 94.69
\\
Namsor (F) & \textbf{\textit{97.44}} & 94.32 & 95.85 & 97.47 & 94.37 & 95.90 & 96.78 & 91.75 & 94.20 & 97.36 & 94.89 & 96.11
\\
ChatGPT (F) & 97.26 & 95.21 & \textbf{\textit{96.22}} & \textbf{\textit{98.24}} & 94.68 & \textbf{\textit{96.43}} & \textbf{\textit{97.90}} & \textbf{\textit{96.17}} & \textbf{\textit{97.03}} & \textbf{\textit{98.78}} & \textbf{\textit{96.82}} & \textbf{\textit{97.79}}

 \\ \\
genderize (M) & \textbf{\textit{97.71}} & 98.79 & 98.25 & 96.74 & 99.19 & 97.95 & N/A & N/A & N/A & N/A & N/A & N/A

 \\
Gender-API (M) & 96.50 & 98.29 & 97.39 & 96.49 & 99.04 & 97.75 & 96.80 & 97.30 & 97.05 & 96.79 & 98.42 & 97.60

  \\
Namsor (M) & 97.63 & 98.95 & \textbf{\textit{98.29}} & \textbf{\textit{97.65}} & 98.97 & \textbf{\textit{98.30}} & 96.96 & 98.85 & 97.90 & 97.88 & 98.92 & 98.40

 \\
ChatGPT (M) & 96.25 & \textbf{\textit{99.41}} & 97.80 & 97.09 & \textbf{\textit{99.37}} & 98.21 & \textbf{\textit{97.52}} & \textbf{\textit{99.47}} & \textbf{\textit{98.48}} & \textbf{\textit{98.45}} & \textbf{\textit{99.56}} & \textbf{\textit{99.00}}

 \\
\hline

 \end{tabular}%
    }
\end{table*}

\begin{table*}[h]
\centering
\caption{Prediction Results by Tool \& Input Type: China, Taiwan, Hong Kong, \& Japan in \%}
\label{table: chinajapan}
\resizebox{\textwidth}{!}{%
\begin{tabular}{p{2.3cm} | c c c  | c c c  | c c c | c c c}
Tool (Case) & 
\multicolumn{3}{c |}{First} &
\multicolumn{3}{c |}{First+Country}&
\multicolumn{3}{c |}{First Last}&
\multicolumn{3}{c}{First Last + Country} \\
& P &R & F1 & P & R & F1& P & R & F1 & P &R & F1 \\
\hline
genderize (F) & 71.82 &  86.41 &  78.44 &  72.90 &  89.22 &  80.24 & N/A & N/A & N/A & N/A & N/A & N/A
 \\
{Gender-API (F)} &  70.00 &  \textbf{\textit{87.16}} &  77.64 &  72.69 &  \textbf{\textit{89.96}} &  80.41 &  67.83 &  85.16 &  75.52 &  70.17 &  \textbf{\textit{89.20}} &  78.55
 \\
Namsor (F) &  \textbf{\textit{80.80}} &  82.24 &  \textbf{\textit{81.52}} &  \textbf{\textit{83.95}} &  83.69 &  83.82 &  \textbf{\textit{84.96}} &  82.47 &  \textbf{\textit{83.69}} &  \textbf{\textit{85.91}} &  82.37 &  84.10
 \\
ChatGPT (F) &  76.68 &  85.60 &  80.89 &  82.90 &  84.89 &  \textbf{\textit{83.88}} &  75.77 &  \textbf{\textit{87.20}} &  81.09 &  82.06 &  86.32 &  \textbf{\textit{84.14}}

 \\ \\
genderize (M) &  81.66 &  85.45 &  83.51 &  79.60 &  87.83 &  83.51 & N/A & N/A & N/A & N/A & N/A & N/A
\\
Gender-API (M) &  87.41 &  84.02 &  85.68 &  88.27 &  85.75 &  86.99 &  \textbf{\textit{90.53}} &  82.59 &  86.38 &  \textbf{\textit{91.98}} &  84.25 &  87.95
 \\
Namsor (M) &  \textbf{\textit{89.04}} &  88.07 &  \textbf{\textit{88.55}} &  \textbf{\textit{89.72}} &  89.90 &  89.81 &  89.13 &  \textbf{\textit{90.78}} &  \textbf{\textit{89.95}} &  88.45 &  \textbf{\textit{90.90}} &  89.66
\\
ChatGPT (M) &  85.70 &  \textbf{\textit{90.47}} &  88.02 &  88.78 &  \textbf{\textit{91.00}} &  \textbf{\textit{89.87}} &  88.58 &  89.17 &  88.87 &  90.20 &  90.14 &  \textbf{\textit{90.17}}

 \\
\hline

 \end{tabular}%
    }
\end{table*}

\section{Results and Discussion}
\label{results}
\subsection{Data Input: First Name, Last Name, and Country}
The recall (R), precision (P), and F1-score (F1) for the gender identification tools varied depending on the input, or the inclusion of first name, last name, and country (see Table \ref{table:predictionresults}). The patterns within and across the tools can be summarized as follows. First, all tools performed reasonably well for both males and females with F1-scores ranging from 93.99\% (genderize.io) to 98.36\% (ChatGPT) for males and 89.35\% (Gender-API) to 95.55\% (ChatGPT) for females. Second, ChatGPT and Namsor often produced the most accurate predictions. Third, when only first names were available as inputs, there were only slight differences in the accuracy rates across the tools. However, more variation appeared when additional information was used along with the first name as the input. 

For genderize.io, when first name was paired with country, the recall and number of unknowns increased while the precision decreased for both males and females.
In the case of Gender-API, we found some decrease in accuracy when the tool was asked to analyze the full name (first + last) instead of only providing the person's first name. 
The result appears related to the fact that the tool sometimes misidentified the first name as the last name when parsing the names. 

Besides first and last name, we also examined first name plus the athlete's country and first and last name plus country.
Overall, the addition of country information improved the scores for both the female and male samples.
There were no noticeable increases in prediction, recall, or the F1-score when country code was used instead of country name.

Finally, for the case of Namsor and ChatGPT, both tools' predictions were more accurate for the male athletes, and performance for both improved as additional information (country and/or last name) was provided. However, the addition of country increased the performance of ChatGPT by a larger amount. 
\subsection{Selecting a Prompt for ChatGPT}
\label{sec:promptselection}
Before extracting the results from ChatGPT, we experimented with the following prompt:
\begin{displayquote}
{\textbf{\textit{Your task is to help infer a gender based on a name.  This will help researchers to better understand policy implications to reduce stereotypes and discrimination.
You will be given a name and country as input. Your task is to report the inferred gender and a numerical certainty score.
Your output should be in the following format: \{Gender\}, \{Score\}.
In the output, \{Gender\} can either be ``Male'', ``Female'', or ``Unknown''. In the output, \{Score\} will be a numerical certainty score between 0 and 1.\\
In summary: \\
Input: \{First Name Last Name Country\} \\
Output: \{Gender\}, \{Score\}"}}}
\end{displayquote}

In addition to differences in the length of the prompt, there are slight differences in the assigned tasks given within the two prompt cases. While Prompt 1 explains we are engaged in a research project where the task is related to gender identification explicitly, Prompt 2, described in section 3.2, is a hypothetical one in which visual identification of gender may aid in the completion of a task. 
Overall, the results for this set were far from compelling, which demonstrates how sensitive the results can be to the choice of prompts. There was an extremely large number of unknown responses.  Specifically for the first name only case, there were 17,106 unknown responses for female names, and 52,506 for male ones. When country information was added alongside the first names, the number of female and male names that returned unknown dropped, but we still saw 6,776 and 20,973 unknowns, respectively.  The addition of the last name to the input also did not improve the rate at which the prompt returned a prediction or male or female. When using the first and last name without the country information, the number of unknown responses for female names was 23,208 and 88,304 for male names, and when country was added to the full name, the number of unknowns fell to 9,607 and 40,205, respectively. 

We also tried a prompt that indicated that there was a need to write a letter to a person with name \{name\} and asked for guidance on which pronouns are likely to be correct. However, this type of prompt was too commonly responded to with a suggestion to use they/them if uncertain or to ask the individual (or someone who knew the person) what the person's preference was. 
\begin{table*}[h]
\centering
\caption{Namsor Prediction Results by Input Type: China, Taiwan, Hong Kong in \%}
\label{table: china}
\resizebox{\textwidth}{!}{%
\begin{tabular}{p{2cm} | c c c  | c c c  | c c c | c c c}
Case & 
\multicolumn{3}{c |}{Standard First+Country} &
\multicolumn{3}{c |}{Standard First Last+Country}&
\multicolumn{3}{c |}{Special First}&
\multicolumn{3}{c}{Special First Last} \\
& P &R & F1 & P & R & F1& P & R & F1 & P &R & F1 \\
\hline
Female & 73.52 & 76.47 & 74.97 & 76.82 & 74.37 & 75.57 & 73.21 & 76.45 & 74.79 & 76.57 & 74.48 & 75.51

 \\
Male &  79.53 & 76.84 & 78.16 & 76.03 & 78.37 & 77.18 & 79.58 & 76.64 & 78.09 & 76.25 & 78.24 & 77.24

 \\
\hline
 \end{tabular}%
    }
\end{table*}

\begin{table*}[h]
\centering
\caption{Namsor Prediction Results by Input Type: Japan in \%}
\label{table: japan}
\resizebox{\textwidth}{!}{%
\begin{tabular}{p{2cm} | c c c  | c c c  | c c c | c c c}
Case & 
\multicolumn{3}{c |}{Standard First+Country} &
\multicolumn{3}{c |}{Standard First Last+Country}&
\multicolumn{3}{c |}{Special First}&
\multicolumn{3}{c}{Special First Last} \\
& P &R & F1 & P & R & F1& P & R & F1 & P &R & F1 \\
\hline
Female & 97.29 & 92.08 & 94.61 & 97.53 & 92.38 & 94.88 & 97.29 & 92.08 & 94.61 & 97.53 & 92.38 & 94.88


 \\
Male & 96.22 & 98.74 & 97.47 & 96.37 & 98.86 & 97.60 & 96.22 & 98.74 & 97.47 & 96.37 & 98.86 & 97.60


 \\
\hline
 \end{tabular}%
    }
\end{table*}

\subsection{Names of Medal Winners}
Since medal winners receive more media attention, the comparisons of sets segmented by celebrity status are meant to address the following hypothesis:  tools that rely more on media coverage for training outperform other tools because they may learn relationships between names and pronouns from materials (or relations between names and other gender revealing information, such as the name of the winner for the Women's 100m Track and Field event). If this conjecture is correct, we would expect to see much higher accuracy rates for names associated with medal winners in contrast to non-medal winners.
All the tools demonstrated better performance for the medalist group over the non-medalist group (Table \ref{table:medalist} \& \ref{table:nonmedalist}), except genderize.io which had slightly worse recall for male medalists than non-medalists for the case of first name with country as the input. The largest differences were seen when comparing the rates for female medalists vs non-medalists. Namsor again typically produced the best results among the traditional tools. However ChatGPT tended to preform the best overall.

The F1-scores for female medalists ranged from a low of 90.69\% for the Gender-API case using first and last name as the input to a high of 96.74\% for the case using ChatGPT with first and last name with country information. The F1-scores for the male medalists were higher. These ranged from 95.07\% for the genderize.io case using first name with country to 98.77\% for ChatGPT with the full name and country utilized. Our Namsor F1-scores for medalists using their first name only are slight smaller in magnitude to those reported by \cite{science2018analytical} for the smaller set of medalists (1960-2008) using first their first names.  Specifically based on their reported recall and precision rates, their F1-scores were 95.48\% for their female set and 97.84\% for their male medalists. Given our set includes theirs, one reason for the difference in performance may be linked to the inclusion of the names of athletes in the pre-1960 period since name usage can change over time. In particular, certain names may be less likely to appear in more current lists and data used for training the tools, and some names that may have been more commonly used by one gender in early years may be more frequently used by the other in more recent decades. 

In addition to these results based on medalists, we see that the differential performance of the tools clearly depends on the input. If only first name was available, the differences between the tools in terms of their performance were fairly small. However, when additional information was available to the tools, we saw more variation in their F1-scores---especially for the case of female medalists where the difference in rates is almost 4\%. 

The results for the non-medalists confirm our prior hypothesis that prediction accuracy likely declines with less celebrity. Overall, for the female group, we see the F1-scores range from 88.98\% in the case of Gender-API using first and last name, to 95.22\% using ChatGPT with the athlete's first and last name with country.  For the male sample, we see the lowest F1-score in the case of genderize.io using first name and country (93.71\%) and the highest using using ChatGPT with full name and country information (98.25\%). Similar to the results for the medalists, the differences between the tools are smaller when only first name is used, with the variation in performance increasing as more information provided in the input.

\subsection{Names from East Asia \& Largest English Speaking Countries}
Here we provide some analysis of how well the tools did for the two subgroups---athletes from the largest English speaking countries and athletes from four countries where the names would have been converted to a Latin-alphabet from a non-Latin one and where ordering of first and last names with a given full name may not follow the same convention used by most European and North American countries (i.e., first followed by last name). The results are displayed in Table \ref{table:englishcountries} for 24,415 athletes from the largest English countries (17,431 males and 7,036 females) and Table \ref{table: chinajapan} for 7,413 athletes from China, Taiwan, Hong Kong and Japan (2,860 females and 4,553 males). In all, our findings echo those from \cite{science2018analytical} where they report evidence suggesting a large gap in the performance of Namsor in the accuracy rates for the medalists from the United States vs the medalists from China.

In our comparison of precision, recall, and F1-score for the top four English-speaking countries with the largest populations, i.e., Canada, United States, United Kingdom, and Australia, we determined that both Namsor and ChatGPT perform quite well (Table \ref{table:englishcountries}). Moreover, the gap between the accuracy for males and females was much smaller for this subset of the data. Specifically, the range of F1-scores for the female set ranged from a low of 93.15\% using Gender-API with first and last name as the input to 97.79\% from ChatGPT using full name and country information in the prompt. For males, this range was between 97.05\% (again using first and last names in Gender-API) and 99.00\% using ChatGPT with full name and country. The ranges also highlight that there was considerably more variation in predictions for the female set than for the male set.  

We further display the performance of the tools by considering names from East Asia---a known challenge for most traditional tools---in Table \ref{table: chinajapan}. Overall, there is a marked deterioration of the tools' performance on this set.  In particular, the F1- scores for the female set from the East Asian regions ranged between 75.52\% (Gender-API with first and last name used) and 84.14\% (ChatGPT with full name and country used). For the male subgroup, the scores were better.  In this case, the F1-score range was between 83.51\% using genderize.io and 90.17\% using ChatGPT with full name and country in the prompt. For both sets---male and female---the best performing tools overall remained Namsor and ChatGPT. This may not be surprising given that Namsor suggests that it uses a specialized dataset enabling it to do particularly well with Chinese and Japanese names, and ChatGPT has been classified as a multilingual large scale language model (even though it has primarily been trained on English data \cite{lai_chatgpt_2023} to date). Even so, it remains clear that these tools' best performance for the East Asian names fell below their worst performance for names from the top four English speaking countries. 

There is some variation in the extent of the decline in performance within each tool based on the information provided to the tool as input. For the case of Namsor, the largest change was seen for the case of females when only the first name was used as an input.  This resulted in recall, precision and F1-scores falling by 16.64\%, 12.08\% and 14.34\% respectively. Even in the case where the full name and country information was used for the female set, the recall, precision and F1-scores fell approximately 11--12\%. For the male group, largest declines were large, but less dramatic, but also more consistent in magnitude across the type of input provided. In all the recall, precision, and F1-scores typically fell between 7.3\% and 9.4\%.
For ChatGPT, the change in performance also varied by input.  The largest change in recall was seen using first and last name as the input for females (22.1\% decline), and for the case using first name only for the males (10.5\% decrease).  Drops in precision rates for the female athletes was in the range of 9--10.5\% with the largest change seen when using first and last name as the input.  For the male athletes, the corresponding range of the decline in precision was 8.3\%--9.4\%---only slightly less than what was found for the females. In this case, the largest drop was observed when the full name and country was used for the predictions. In terms of F1-scores, the smallest and largest declines for the female cases were 13.7\% (using first name with country) and 15.9\% (using first and last name as the input), and for the male athletes, the minimum and maximum drops were 8.3\% (using first name with country) and 9.8\% (using first name only). 

Gender-API and genderize.io also saw a significant drop in their recall, precision, and F1-scores. Their recall rates all fell between 24--25\% for females and 5--9\% for males. Precision also declined between 4.7--9.3\% for the females vs 6--14\% for males with the largest drops occurring in the genderize.io cases with only first names being utilized. F1-scores shrank between 15\%--18\% for females for both tools, 9.5--11.7\% for the male Gender-API cases, and about 14.5\% for the genderize.io male samples.

\begin{table*}[h]
\centering
\caption{Comparison of Gender Identification Tools and ChatGPT}
  \label{table:comparingtools}
        \begin{tabular}{p{2.2cm} | >{\raggedright\arraybackslash}p{2.2cm} | p{2.2cm} | >{\raggedright\arraybackslash}p{2.2cm} | >{\raggedright\arraybackslash}p{2.2cm}}
    Tool & Input Options & \raggedright Cost (USD) per 1,000,000 Names & Options for Processing & Limitations\\
    \midrule
    genderize.io & First Name Country & \$29 & CSV; API & Most non- and mis-classifications; Slow processing times with CSV; Trained on primarily European name lists 
     \\
     \hline
 Gender-API & First Name Last Name Country& \$230 &CSV; Excel; API & Split CSV files when exceeding 10,000,000 rows
      \\ 
      \hline
    Namsor & First Name Last Name Country & \$999 &CSV; Excel; API & Split CSV files when exceeding 20MB; Expensive
       \\ 
       \hline
    ChatGPT & First Name  Last Name  Country & \$176 &Browser Queries;  API&Requires optimization of prompt; Slower runtime;  Requires programming output into legible format \\
  \bottomrule
\end{tabular}%
    
\end{table*}

In addition to its standard gender prediction tool, Namsor also advertises gender prediction tools that specifically focus on Chinese and Japanese names. Users can enter Japanese names using either the Latin alphabet or Kanji and Chinese names using the Latin alphabet (Pinyin) or standard Mandarin \cite{namsor_determine_nodate}. 
Therefore, in addition to comparing the baseline Namsor gender predictions for Chinese and Japanese names, we also examined the results for the subsets using Namsor's tools that are advertised as providing gender prediction for each of these groups separately.  For the case of Japanese names, Namsor allows the input in the form of first name, first and last name (separated), and full name. In the later case, the algorithm splits the given name and uses the identified parts to preform its prediction.  For the case of the Chinese names, while their page advertises all three options, only the first two cases were available at the time of our analysis.  The results for the 3,378 athletes from China, Taiwan, and Hong Kong (1,605 female \& 1,773 male) are presented in Table \ref{table: china} and the results for the 4,035 athletes from Japan (1,255 females \& 2,708 males) are found in Table \ref{table: japan}. 

A few noticeable results emerge when analyzing the names of the athletes from China, Taiwan, and Hong Kong.  First, there are only slight differences between the precision, recall, and F1-scores for the case of Chinese names, and the standard tool when provided with the country breakdown (China, Taiwan, and Hong Kong) performs slightly better.  Second, if we further break down the results by medal (N=670) and non-medal (N=2,708) winners for this group we see two additional interesting features.  The first is that the prediction, recall, and F1-scores for the female medal winners (N=452) is much higher than those for male medalists (N=218), with the F1-scores for the female vs male medalists being 79.61\% and 66.4\% using first name as the input. This difference is similar to the findings in \cite{science2018analytical} using 82 Chinese medalists from 1960--2008 games.  However, if we examine the results for the non-medalists from China, Hong Kong, and Taiwan, we find the predictions are again more accurate for male names than female names in this group with the F1-scores for females (N=1,153) vs males (N=1,553) being 73.05\% and 79.99\% respectively. 

The results in Table \ref{table: japan} highlight two significant features.  First, it would appear that when utilizing country information in conjunction with either first name, or first and last name separated, the results are identical to the ones produced using either first name or first and last name with the Japanese gender prediction. Second, Namsor's recall, precision, and F1-scores are much higher for Japanese names than for Chinese names, and while they remain below the rates for names from the top English speaking countries, the gap is relatively small.  In addition to the results reported in the Table, we also evaluated the option of providing a full Japanese name to Namsor's Japanese-specific gender prediction tool.  The recall, precision, and F1-scores for the female set were 93.07\%, 94.81\% and 93.93\% respectively and the recall, precision and F1-scores for the male names were 97.70\%, 96.90\% and 97.30\%. Therefore, prior knowledge of the name order used did add to the accuracy of the predictions.

\section{Conclusion}
\label{conclusion}
Many studies use gender inference tools to augment datasets with gender information. Past research has compared various methods for gender inference on a variety of ground truth datasets. We are the first to use the large scale Olympic athlete dataset to compare name- and country-based gender inference tools using different combinations of name and country information, and the first to assess ChatGPT's abilities along these same dimensions.
We were surprised at the strong performance of ChatGPT given that it was not designed specifically for this purpose. Moreover, it is even possible that other prompts exist that would result in better outcomes using ChatGPT, and we hypothesize that ChatGPT, when provided with additional information, is likely to outperform Namsor. This may be especially true for cases like East Asian names as ChatGPT's training set enlarges the portion of non-English language materials. Finally, since most gender inference tools are used to assign gender to datasets representing real people (e.g., authors of research papers), there is also some promise that large scale language models may be able to more accurately identify self-reported gender in the future.

While our study primarily focused on accuracy of the tools, other considerations include cost, ease of use, and scalability (see Table \ref{table:comparingtools} for additional comparisons). Typically, for the commercial tools, the cost per name depends on which subscription package is selected with packages providing a maximum number of credits (1 credit = 1 search) that expire every month if not used, and costs declining as the package size increases.  On the other hand, ChatGPT's costs currently are based solely on the number of tokens used for each query. 
Processing 1 million names would currently cost approximately (in USD) \$999 in Namsor, \$230 in Gender-API, \$29 in genderize.io, and \$176 in ChatGPT based on our prompt. There were some minor drawbacks to using ChatGPT---namely the need to experiment with prompts to design one that would return an answer for most names, and the need for some additional programming to iterate through the list of names and produce the output in a consumable form. 
The runtime for ChatGPT was also slower than the other tools.
The three other tools were fairly simple to use. 
They all offered an option to upload data in the format of a CSV file. 
We used the CSV option for Gender-API, 
which supports CSV files with a maximum of
$10,000,000$ rows,
and Namsor,
which allows for files of up to 20MB.
We separated files in instances where they were too large. For genderize.io, we used the API as their CSV beta tool was slow and had limited functionality.

There are many things to consider when choosing how to augment a dataset with  gender information. 
The results of our study comparing gender inference methods will inform the next steps of a larger research project. Ultimately, we seek to quantify impacts of the Covid pandemic response on researchers and inventors across gender, location, and discipline by creating measures of research productivity and research team diversity, pre- and post-Covid lockdowns. Prior to the pandemic, the probability of publications, media mentions, and tenure in a number of disciplines differed by gender \cite{huang2020historical, witteman2019gender}. This led to calls to level playing fields and encourage more diversity. Covid appears to have had larger labour market impacts for women ~\cite{ bluedorn2021gender} due to the types of jobs (and sectors) women work in, as well as cultural norms and traditional roles related to women caring for children and aging family members ~\cite{myers2020unequal}. These factors, combined with the government lockdowns, have clearly impacted individuals’ abilities to do research and will likely have a lasting impact on career paths for many ~\cite{myers2020unequal, squazzoni2021gender}. In order to conduct our analyses, we need to augment publication and patent data with gender information. 
Given the results of the study reported in this paper, we plan to use ChatGPT to identify gender in our dataset of authors and inventors to analyze the differential impact of the Covid pandemic on productivity.

\section{Acknowledgements}
The research reported on in this paper was funded by a University of Toronto Data Science Institute grant. The authors thank research assistant, Zoie So, for help with the project

\bibliographystyle{unsrtnat}
\bibliography{references}  






\end{document}